\documentclass[conference]{llncs}
\usepackage{cite}
\usepackage{amsmath,amssymb,amsfonts}
\usepackage{algorithmic}
\usepackage{graphicx}
\usepackage{subcaption}
\usepackage{textcomp}
\usepackage{xcolor}
\usepackage{tabularx}
\usepackage{threeparttable}
\usepackage{booktabs}
\def\BibTeX{{\rm B\kern-.05em{\sc i\kern-.025em b}\kern-.08em
		T\kern-.1667em\lower.7ex\hbox{E}\kern-.125emX}}
\begin{document}
	
	\title{Affine-Transformation-Invariant Image Classification by Differentiable Arithmetic Distribution Module}
	
	\author{Zijie Tan \and Guanfang Dong \and Chenqiu Zhao \and Anup Basu}
	\institute{Multimedia Research Center, University of Alberta, Canada \\
		\email{ztan4@ualberta.ca, guanfang@ualberta.ca, chenqiu1@ualberta.ca, basu@ualberta.ca}}

	\maketitle
	
	\begin{abstract}
		Although Convolutional Neural Networks (CNNs) have achieved promising results in image classification, they still are vulnerable to affine transformations including rotation, translation, flip and shuffle. 
		The drawback motivates us to design a module which can alleviate the impact from different affine transformations. 
		Thus, in this work, we introduce a more robust substitute by incorporating distribution learning techniques, focusing particularly on learning the spatial distribution information of pixels in images. 
		To rectify the issue of non-differentiability of prior distribution learning methods that rely on traditional histograms, we adopt the Kernel Density Estimation (KDE) to formulate differentiable histograms. 
		On this foundation, we present a novel Differentiable Arithmetic Distribution Module (DADM), which is designed to extract the intrinsic probability distributions from images. 
		The proposed approach is able to enhance the model's robustness to affine transformations without sacrificing its feature extraction capabilities, thus bridging the gap between traditional CNNs and distribution-based learning. 
		We validate the effectiveness of the proposed approach through  ablation study and comparative experiments with LeNet.
	\end{abstract}
	
	\vspace{-5pt}
	\section{Introduction}	
	Convolutional Neural Network (CNN) has been a powerful and popular tool for extracting features from images, enabling state-of-the-art performance in various computer vision tasks, such as object detection, image segmentation, and image classification \cite{wang2019development, ren2015faster, du2020medical, lei2019dilated}. However, CNNs are inherently weak against some simple affine transformations such as rotation, since they rely strongly on the spatial patterns in data \cite{zhao122017marginalized}. 
	Compared to the pattern information, distribution information has the potential to provide affine-transform invariance. 
	Because it concentrates on the overall statistical properties and probability distribution of pixels regardless of their exact spatial arrangement. 
	Therefore, to alleviate the aforementioned issue, we propose to learn the distribution information for classification task, in which the distribution learning techniques are incorporated. 
	
	Distribution learning techniques are a group of methods wherein the focus is shifted from learning explicit patterns to understanding the underlying statistical distributions and characteristics of the data. These approaches often seek to capture the broader, holistic properties of datasets rather than narrowly focusing on specific, local patterns. For example, DIDL network \cite{dong2023learning} utilizes the temporal distribution of pixel values across video frames, and learns the underlying statistical features of background and foreground in different scenes. In this work, we will instead focus on learning the spatial distribution information of pixels in images, which can enable our model to capture the affine-transformation-invariant features in input data, and thus make it more robust against transformations such as rotation. 
	
	A common approach to describe distribution in computer vision is the histogram of pixel values. 
	However, the construction of histograms is not differentiable, which makes it hard to efficiently integrate histograms and neural networks. To address this problem, we utilize the Kernel Density Estimation (KDE) to approximate the histogram. The key idea is that rather than counting pixels and assigning them to discrete bins, we represent the data distribution by overlaying a kernel function at each data point and summing the contributions of these kernels across all data points. In this way, the bins of histograms are transformed into smoothed probability density instead of discrete counts, while preserving the statistical information. 
	
	In this work, a KDE-based method is formulated for constructing differentiable histograms from images. 
	Based on this, we propose a novel differentiable arithmetic distribution module, which is explicitly crafted to learn the underlying probability distribution of the input space. 
	This global feature fortifies the model's robustness against specific affine distortions, notably rotations. 
	At the same time, the differentiability enables spatial feature extraction for the distribution learning techniques.
	The main contributions of this work are summarized below:
	\begin{itemize}
		\item We propose the Differentiable Arithmetic Distribution Module (DADM) that is adept at extracting inherent distribution information from images while also offering resilience to certain affine transformations, such as rotations.
		\item We utilize a KDE-based approach of constructing smoothed and differentiable histograms, enabling a seamless integration of histograms and neural network. 
		\item We conduct experiments to evaluate and demonstrate the effectiveness and robustness of the formulated method, including comparison with the famous CNN-variant LeNet and an ablation study of the proposed DADM network. 
	\end{itemize}
	
	\section{Related Work}
	Distribution learning is a technique that focuses on understanding the underlying probability distributions of data from the observed samples, rather than solely identifying explicit patterns or features \cite{kearns1994learnability}. Modelling the entire data distribution can provide a holistic view, which enhances invariance to common transformations such as translations, scalings, and rotations. In addition, the distribution information has been shown to be instrumental in many tasks in computer vision, including image generation, background subtraction, segmentation \cite{zhao2023learning}. 
	\par
	Although a histogram is an appropriate way to describe the probability density function and thus to convey the distribution information, traditional histogram is discrete and non-differentiable, making it challenging to be directly integrated into modern neural network frameworks that rely on gradient-based optimization and backpropagation \cite{peeples2022histogram}. To address this issue, multiple methods have been proposed to approximate the soft histogram in a continuous and differentiable manner. One notable approach is the HOG \cite{chiu2015see} which utilizes linear filtering operations and convolutions to approximate a piecewise differentiable histogram for pose estimation. Wang et al. proposed the first learnable histogram layer for neural networks by formulating HOG with a series of convolutional modules \cite{wang2016learnable}. Furthermore, Sedighi et al. presented a globally differentiable histogram layer by utilizing radial basis functions as step functions in the backpropagation \cite{sedighi2017histogram}. Peeples et al. further extended their work to have adaptive number and width of bins \cite{peeples2021histogram}. 
	\par
	While most of the aforementioned methods assume a predefined distribution of the data, KDE offers the flexibility to estimate the underlying distribution directly from the samples without prior assumptions \cite{elgammal2002background}. For example, though not directly applying KDE, HistoGAN \cite{afifi2021histogan} uses an inverse-quadratic kernel function to compute a weighted contribution of pixels to the bins of output histogram. 
	Another related method is DeepHist \cite{avi2020deephist}, which is also based on KDE. 
	It uses a sigmoid-based kernel function to estimate the histogram of pixels to separate the edge and color features in images. 
	Nevertheless, this method does not thoroughly explore the application of probability and distribution information, but primarily use the histogram as a way to represent color information.
	Our method, on the other hand, utilizes the Gaussian kernel and considers the histogram in a probabilistic viewpoint, which adds values to the robustness and effectiveness in image classification. 
	
	\par
	Based on these explorations, researchers seek to optimize the integration of histograms and neural networks. For example, ImHistNet \cite{hussain2021learnable} is capable of learning complex and subtle task-specific textural features and global statistical
	features directly from the image intensity, which is defined by a set of convolution operations. Similarly, PTFEM \cite{zhu2021learning} is a texture enhancement network module that uses an adaptive histogram equalization mechanism that pays specific attention to texture details and propagates the distribution information across pyramid layers. While these traditional neural networks only use convolution operations to handle distribution information, the DIDL network \cite{dong2023learning} presents the arithmetic distribution layers that directly consider histograms as probability density functions. 
	However, the construction of histogram in a DIDL network is not differentiable, limiting its feature extraction ability and computation speed. In this work, we enhance the arithmetic distribution modules in the DIDL network with the proposed KDE-based differentiable histogram module. 
	
\begin{figure*}[!htbp]
	\centering
	\includegraphics[width=\textwidth]{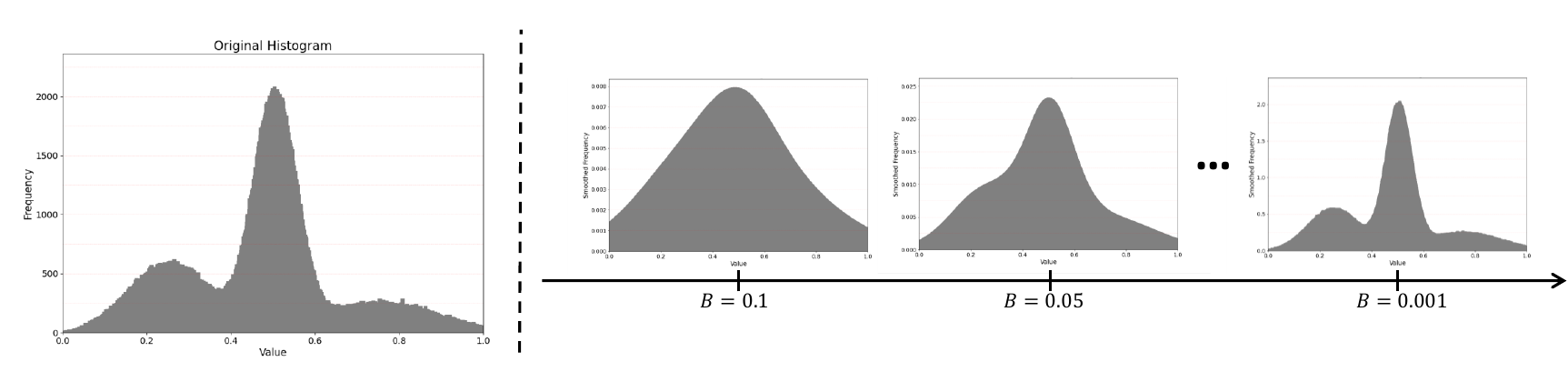}
	\caption{\centering An example of KDE-based approximations of histogram with different bandwidth \(B\). Note that the smoothed histogram gets closer to the original histogram as \(B\) grows. We choose \(B=0.001\) in our implementation. }
	\label{diff_hist}
\end{figure*}
	\section{Method}
	In this section, we will discuss the mathematical details and implementations of the proposed differentiable histogram module and the corresponding DADM network. Since we focus on the task of image classification, we can first assume that the input space is the set of gray-scale images with one single channel ranging from 0 to 255. This will not compromise the generalizability of our approach because each channel of multi-channel images can also be seen as a gray-scale image. In our implementation, the pixel values are narrowed into the range of \([-1, 1]\). 
	\subsection{Differentiable Histogram}
	We represent the distribution of pixels in image with histogram since it is simple and straightforward to interpret. And more importantly, it does not assume a specific distribution for the data. This can be advantageous when dealing with data of an unknown or complex distribution \cite{kontkanen2007mdl}.
	To integrate histogram into the arithmetic distribution model, we approximate it using differentiable functions, as illustrated in Fig~\ref{diff_hist}. 
	\par
	To transform the discrete histogram into a smooth, differentiable approximated representation, we will first need to partition the range of pixel values. The details of partitioning and the notations for deriving the differentiable histogram are given below.
	\par
	\textbf{Notation:} Let \(N\) be the number of bins in the desired representation of histogram, where each bin will have the width \(W=\frac{1-(-1)}{N}=\frac{2}{N}\) and span \(\Delta=\frac{W}{2}\). For the i-th bin, it has left bound \(L_i=-1+(i-1)W\) and right bound \(R_i=-1+iW\). Therefore, the i-th bin is \(X_i=[-1+(i-1)W, -1+iW]\) centering at \(\mu_i=-1+(i-\frac{1}{2})W\), where \(i=1, 2, 3, ..., N\).  
	\par
	Inspired by DeepHist \cite{avi2020deephist}, we also utilize the KDE to compute the differentiable approximation of the histogram, which operates by placing a kernel on each data point (or pixel value in our case) and summing up the contributions from all these kernels to obtain a smooth probability density function. Given a set of pixel values \(x_1, x_2, ..., x_M\) from image, the KDE estimate at point \(x\) in the input space is given by:
	\begin{equation} \label{density}
		\hat{f}(x) = \frac{1}{MB}\sum_{j=1}^M \mathcal{K}(\frac{x-x_j}{B}), 
	\end{equation}
	where \(\mathcal{K}(\cdot)\) is the selected non-negative kernel function, and \(B>0\) is a free parameter called bandwidth which controls the smoothness of the estimated function. And then by definition, the probability of a bin \(X_i\) is given by the integral of the KDE function over the bin's range:
	\begin{equation} \label{prob}
		P(X_i=\mu_i) = \int_{L_i}^{R_i} \hat{f}(x)dx = \int_{\mu_i-\Delta}^{\mu_i+\Delta} \hat{f}(x)dx
	\end{equation}
	\par
	Combining the Equations \ref{density} and \ref{prob}, we derive a unified expression for computing the probability of bins with respect to the kernel function: 
	\begin{align}\label{eq_w_KDE}
		P(X_i=\mu_i) &=\int_{\mu_i-\Delta}^{\mu_i+\Delta} \hat{f}(x)dx \notag \\ 
		&=\int_{\mu_i-\Delta}^{\mu_i+\Delta} \frac{1}{MB}\sum_{j=1}^M \mathcal{K}(\frac{x-x_j}{B})dx \notag \\
		&=\frac{1}{MB}\sum_{j=1}^M \left[\int_{\mu_i-\Delta}^{\mu_i+\Delta} \mathcal{K}(\frac{x-x_j}{B})dx\right]
	\end{align}
	\par
	The actual value of the above equation will depend on the selection of kernel function \(\mathcal{K}(\cdot)\). Some popular kernel functions in KDE include Gaussian Kernel, Epanechnikov kernel, Uniform Kernel, etc. In this paper, we choose the Gaussian distribution \(\mathcal{K}(x)=\frac{1}{\sqrt{2\pi}}e^{-\frac{x^2}{2}}\) as the kernel function. 
	The final equations for computing our differentiable histogram are given by:
	\begin{align} \label{eq_gaussian}
		P(X_i=\mu_i)
		&=\frac{1}{MB}\sum_{j=1}^M \left[\int_{\mu_i-\Delta}^{\mu_i+\Delta} \mathcal{K}_1(\frac{x-x_j}{B})dx\right] \notag\\
		&=\frac{1}{MB}\sum_{j=1}^M \left[ erf(\frac{x-x_j}{B})\bigg|_{\mu_i-\Delta}^{\mu_i+\Delta}\right] \notag\\
		&=\frac{1}{MB}\sum_{j=1}^M \mathcal{G}_i(x_j), \\
		\text{where } \mathcal{G}_i(x) &= erf(\frac{\mu_i-x+\Delta}{B}) - erf(\frac{\mu_i-x-\Delta}{B}) \notag\\
		\text{and }  erf(x)&=\frac{2}{\sqrt[]{\pi}}\int_0^x e^{-t^2}dt. \notag
	\end{align}
	%

	\subsection{Differentiable Arithmetic Distribution Learning}
	\begin{figure*}[ht]
		\centering
		\includegraphics[width=\textwidth]{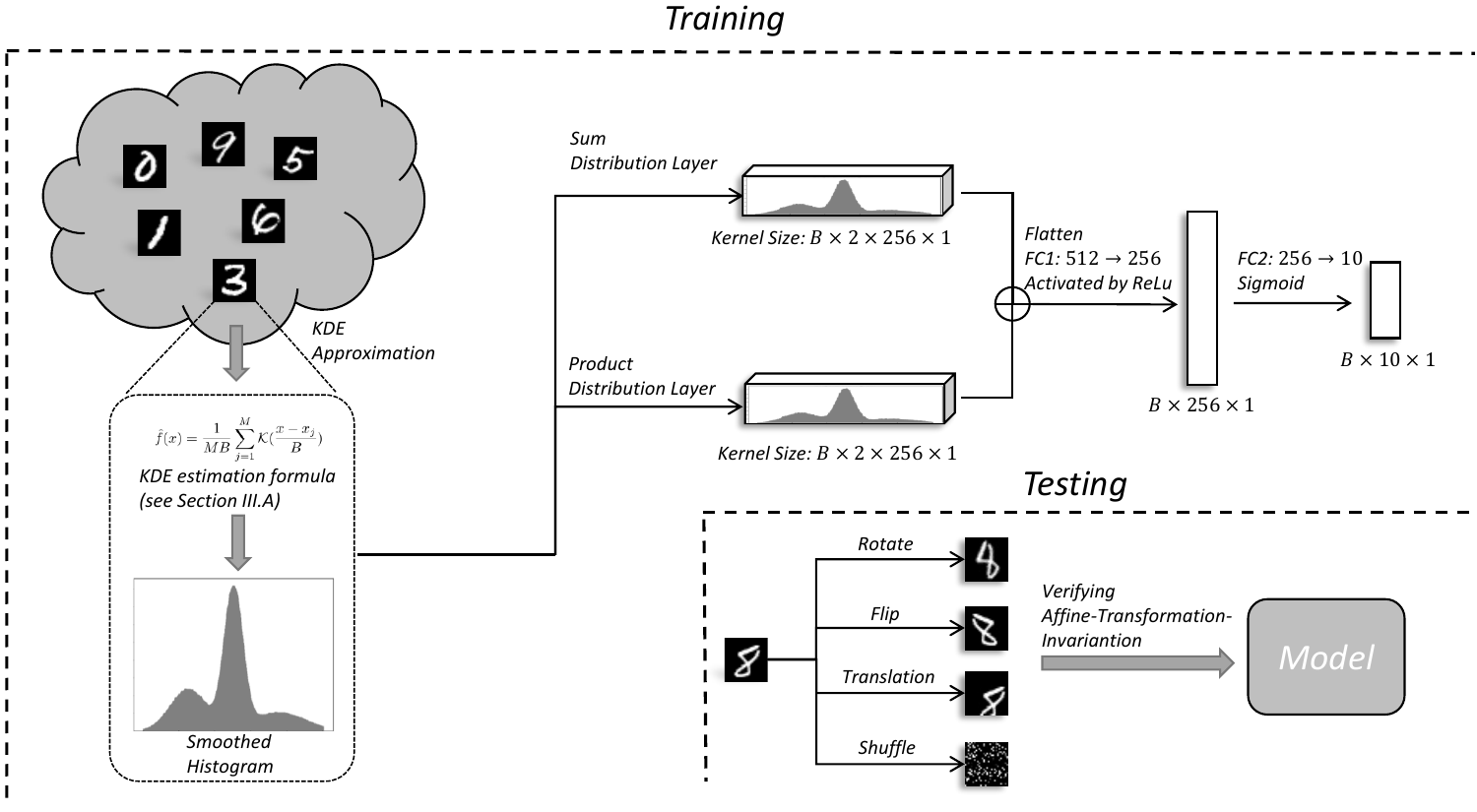}
		\caption{Network architecture and pipeline for our proposed method. First, an image is transformed into a smoothed histogram via Kernel Density Estimation (KDE). Then, the smoothed histogram is processed by the Sum Distribution Layer and Product Distribution Layer to learn distributional information. Each layer incorporates two $256 \times 1$ histogram kernels. Finally, the outputs from both layers are classified through a Fully Connected Layer. During testing, to verify affine transformation invariance, the input images are rotated, flipped, translated, and shuffled. These affine transformations are applied only during the testing phase.}
		\label{process}
	\end{figure*}
	Together with the obtained differentiable function above, we adapt the product distribution layer and sum distribution layer proposed in our previous work \cite{dong2023learning} to design the differentiable arithmetic distribution module. For understandability, we will also introduce these two layers below. 
	\par
	In contrast to the convolution layer, which view input histograms merely as vectors, the product distribution layer and sum distribution layer interpret them as distributions to better describe the probability information and the correlation between histogram entries. To accomplish this, these two layers represent their learning kernels as histograms. That is, given the input and output distributions denoted by random variables \(X\) and \(Z\), the distribution layers learn the distributions of learning kernels denoted by random variables \(W\) and \(B\) such that \(Z=WX+B\). Note that all distributions are described by histograms in our work. 
	The expressions of forward pass and backpropagation are formulated as follow: 
	\par
	Product distribution layer:
	\begin{equation}
		\begin{aligned}
			f_Z(z)&=\int_{-\infty}^{\infty} f_W(w)f_X(\frac{z}{w})\frac{1}{|w|}dw ,\quad\text{forward}\\
			\nabla w_i&=\sum_{j=-\infty}^{\infty}\nabla z_j f_X(\frac{z_j}{i})\frac{1}{|i|},\quad\text{backward}
		\end{aligned}
	\end{equation}
	where \(f_Z(z)\), \(f_W(w)\), and \(f_X(x)\) are probability density functions (PDF) that represent the distributions of \(Z\), \(W\), \(X\), respectively. And \(z\), \(w\), and \(x\) are the entries of corresponding histogram. 
	\par
	Sum distribution layer:
	\begin{equation}
		\begin{aligned}
			f_Z(z)&=\int_{-\infty}^{\infty}f_B(b)f_X(z-b)db, \quad\text{forward}\\
			\nabla b_k&=\sum_{j=-\infty}^{\infty}\nabla z_j f_X(z_j-k), \quad\text{backward}
		\end{aligned}
	\end{equation}
	where, likewise, \(f_B(b)\) and \(b\) are the PDF and histogram entries of the distribution represented by \(B\), respectively. 
	\par
	The proposed neural network module consists of a differentiable histogram layer, a product distribution layer, and a sum distribution layer. 
	Specifically, the input images of size \(B \times 1 \times H \times W\) will be first fed into the differentiable histogram layer to generate the smoothed histograms of size \(B \times N \times 1\), where \(B\) is the batch size and \(N\) is a parameter representing the number of bins in the desired histogram. The distribution layer then employs the learning kernel of size \(N \times 1\) on this histogram and generate the output of the same shape. This output will be further fed into the classifier module for image classification result. 
	More details of network architecture and the pipeline are show in Fig.~\ref{process}.  
	The parameter configuration of network architecture is given in Tab~\ref{network_details}.
	
	\section{Experimental Results}
	\begin{table*}[htbp]
		\centering
		\begin{threeparttable}
			\caption{Details of architecture of the proposed model and the networks selected for experiments.}\label{network_details}
			\begin{tabularx}{\textwidth}{|l|>{\centering\arraybackslash}X|l|>{\centering\arraybackslash}X|l|>{\centering\arraybackslash}X|l|>{\centering\arraybackslash}X|l|>{\centering\arraybackslash}X|}
				\hline
				\multicolumn{2}{|c|}{\textbf{LeNet}} & \multicolumn{2}{c|}{\textbf{Base Classifier}} & \multicolumn{2}{c|}{\textbf{CNN}} & \multicolumn{2}{c|}{\textbf{DADM}} \\
				\hline
				type & size & type & size & type & size & type & size \\
				\hline
				Conv & $(1, 5, 5) \times 6$ & & & & & & \\
				Relu &  & & & & & & \\
				MaxPool & \((2, 2)\) & & & & & & \\
				Conv & \((6, 5, 5) \times 16\) & & & & & & \\
				Relu &  & & & Conv & \((1, 3, 3) \times 4\) & DiffDis & \\
				MaxPool & \((2, 2)\) & & & Relu &  & ProDis & \(1 \times 256 \times 1\) \\
				Linear & \(120 \times 256\) & Linear & \(256 \times 784\) & Linear & \(256 \times 784\) & SumDis & \(1 \times 256 \times 1\) \\
				Relu &  & Relu &  & Relu &  & Relu &  \\
				Linear & \(84 \times 120\) & Linear & \(512 \times 256\) & Linear & \(512 \times 256\) & Linear & \(512 \times 256\) \\
				Relu &  & Relu &  & Relu &  & Relu &  \\
				Linear & \(10 \times 84\) & Linear & \(10 \times 512\) & Linear & \(10 \times 512\) & Linear & \(10 \times 512\) \\
				Softmax &  & Softmax &  & Softmax &  & Softmax &  \\
				\hline
			\end{tabularx}
			\begin{tablenotes}
				\small
				\item Note: Conv - Convolution layer, Relu - Rectified Linear Unit, ProDis - Product Distribution Layer, SumDis - Sum Distribution Layer, DiffDis - Differentiable Histogram Layer.
			\end{tablenotes}
		\end{threeparttable}
	\end{table*}
	\begin{figure*}[ht]
		\centering
		\includegraphics[width=\textwidth]{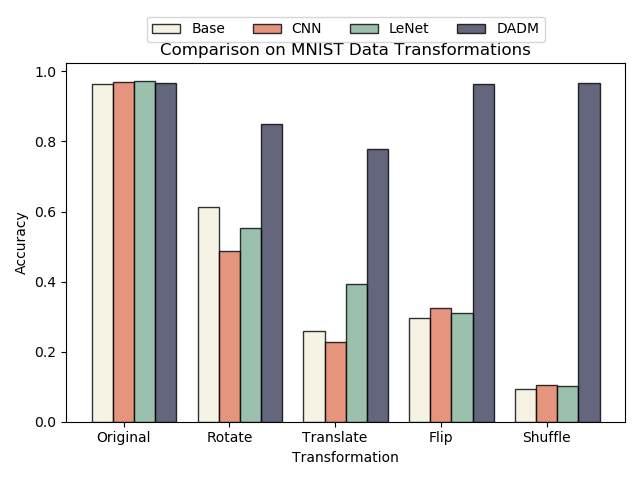}
		\caption{\centering Histogram showing the Top-1 accuracy (\%) of baseline methods and DADM on MNIST dataset under various transformations. }
		\label{bar_mnist}
	\end{figure*}
	In this section, we will discuss the experiments we conducted to evaluate the performance of method and examine the affine transformation invariance. All training and testing are performed on a NVIDIA RTX A4000. During training, we applied the Adam optimizer with a learning rate of 0.001 and the Negative Log Likelihood function as the loss function. An overall illustration of the evaluation results is shown in Figure~\ref{bar_mnist}.

	\subsection{Dataset}
	To validate the robustness of our proposed model, we employ the MNIST dataset, which is a widely-recognized benchmark in computer vision research. The dataset encompasses 60,000 training images and an additional 10,000 testing images, each featuring handwritten digits ranging from 0 to 9. All images are in grayscale and have dimensions of 28$\times$28 pixels. We subject the MNIST images to various transformations in order to evaluate and compare the models' performance under varying conditions.
	
	\begin{table}[htbp] 
		\caption{Results of experiments comparing LeNet and our method on the original and transformed data in MNIST, reported in Top-1 accuracy (\%). Numbers in parentheses are the performance decrease compared to tests on original images, where lower values represent better performance.}
		\label{ccnn}
		\begin{threeparttable}
			\begin{tabularx}{\linewidth}{p{1.35cm}|p{1.5cm}|X|X|X|X}
				\toprule
				& \multicolumn{1}{c|}{Original} & \multicolumn{1}{c|}{Rotate} & \multicolumn{1}{c|}{Translate} & \multicolumn{1}{c|}{Flip} & \multicolumn{1}{c}{Shuffle} \\
				
				\hline
				LeNet & 97.39 & 55.20 (42.19) & 39.30 (58.09) & 31.09 (66.3) & 10.33 (87.06) \\
				DADM & 96.57 & \textbf{85.02}(\textbf{11.65}) & \textbf{77.93}(\textbf{18.74}) & \textbf{96.34}(\textbf{0.33}) & \textbf{96.56}(\textbf{0.11}) \\
				\bottomrule
			\end{tabularx}
		\end{threeparttable}
	\end{table}		
	\subsection{Comparison with Convolutional Neural Network} \label{cwcnn}
	To highlight the advantages of our proposed methodology over traditional Convolutional Neural Networks, we conduct a comprehensive comparative analysis against the well-known LeNet \cite{lecun1998gradient} architecture. The details of architecture and parameter are given in Table~\ref{network_details}. Specifically, we evaluate the model's robustness and performance under various affine transformations. The transformations in experiments include rotation, translation, and flipping. In addition, we also test the model's performance on randomly shuffled images. Training is performed on the training set of the original MNIST dataset, while the testing are performed on the testing set of the original and transformed dataset. 
	\par
	The details of selected transformation in experiment are listed as follow: for the rotation, input images are rotated by a random degree ranging from 0 to 90; for the translation, input images are shifted to a arbitrary direction for a random number of pixels with a maximum offset of 8; for the flipping, input images are randomly flipped horizontally or vertically. 
	\par
	The results of experiments are listed in Table~\ref{ccnn}. Through the result of comparison between LeNet and our method, we can see that though both method perform closely on the original MNIST dataset with our method being only slightly less favorable, our method significantly outperforms LeNet across all other categories of experiments where specific transformations are applied to the input images. This demonstrates that our method is much more robust against affine transformations than the classical CNN architectures such as LeNet, while maintaining a comparable power to them in terms of effectiveness.

	\begin{table}[htbp] 
		\caption{Results of experiments comparing the base classifier, CNN, and our method on the original and transformed data in MNIST, reported in Top-1 accuracy (\%). Numbers in parentheses are the performance decrease compared to tests on original images, where lower values represent better performance.} 
		\label{ablation_results}
		\begin{threeparttable}
			\begin{tabularx}{\linewidth}{p{1.35cm}|p{1.5cm}|X|X|X|X}
				\toprule
				& \multicolumn{1}{c|}{Original} & \multicolumn{1}{c|}{Rotate} & \multicolumn{1}{c|}{Translate} & \multicolumn{1}{c|}{Flip} & \multicolumn{1}{c}{Shuffle} \\
				
				\hline
				Base & 96.29 & 61.23 (35.06) & 25.97 (70.32) & 29.69 (66.60) & \hspace{4.5pt}9.39 (86.90) \\
				CNN & 97.03 & 48.82 (48.21) & 22.65 (74.38) & 32.37 (64.66) & 10.43 (86.60) \\
				DADM & 96.57 & \textbf{85.02}(\textbf{11.65}) & \textbf{77.93}(\textbf{18.74}) & \textbf{96.34}(\textbf{0.33}) & \textbf{96.56}(\textbf{0.11}) \\
				\bottomrule
			\end{tabularx}
		\end{threeparttable}
	\end{table}		
	\subsection{Ablation Study}
	We perform an ablation study to further demonstrate the contributions of individual components in our proposed neural network architecture. Specifically, we investigate the role of DADM. As shown in Table~\ref{network_details}, we derive two baseline models for comparative analysis by removing the differentiable arithmetic distribution module and replacing it with a convolution module. They are tested under a similar experiment configuration to the one in Section~\ref{cwcnn}, where various transformations are applied to input data. Likewise, the training only use the training images of the original MNIST dataset, and the testing use the original and transformed testing images. 
	\par
	The results are summarized in Table~\ref{ablation_results}. Several observations can be made from the results. For the original data, the extremely narrow gap between our method and the two baselines suggests that the proposed approach maintains a similar power as the CNN in terms of effectiveness. While for the transformed data, our method again demonstrates its robustness against the affine transformations since its accuracy are significantly higher than that of the baselines.

	\subsection{Case Study}
	\begin{table}[htbp] 
		\caption{The class-wise classification performance of LeNet and DADM on MNIST, reported in Top-1 accuracy (\%). }
		\label{case_study}
		\begin{threeparttable}
			\begin{tabularx}{\linewidth}{p{1.1cm}|X|X|X|X|X|X|X|X|X|X}
				\toprule
				& \multicolumn{2}{c|}{Original} & \multicolumn{2}{c|}{Rotate} & \multicolumn{2}{c|}{Translate} & \multicolumn{2}{c|}{Flip} & 
				\multicolumn{2}{c}{Shuffle} 
				\\
				& \multicolumn{1}{c|}{LeNet} & \multicolumn{1}{c|}{DADM} & \multicolumn{1}{c|}{LeNet} & \multicolumn{1}{c|}{DADM} & \multicolumn{1}{c|}{LeNet} & \multicolumn{1}{c|}{DADM} & \multicolumn{1}{c|}{LeNet} & \multicolumn{1}{c|}{DADM} & \multicolumn{1}{c|}{LeNet} & \multicolumn{1}{c|}{DADM}  
				\\
				\hline
				Class 0 & 98.44 & 96.07 & 87.77 & 91.90 & 26.21 & 69.09 & 59.55 & 95.84 & 3.14 & 96.07  
				\\
				Class 1 & 98.40 & 99.43 & 63.38 & 96.22 & 35.34 & 98.41 & 81.49 & 99.20 & 6.33 & 99.43
				\\
				Class 2 & 96.73 & 95.05 & 55.12 & 80.37 & 44.07 & 70.28 & 6.28 & 94.82 & 20.15 & 95.05
				\\			
				Class 3 & 97.44 & 96.59 & 47.99 & 79.36 & 48.34 & 72.50 & 38.72 & 96.35 & 21.05 & 96.59
				\\
				Class 4 & 97.54 & 96.74 & 53.79 & 85.84 & 38.44 & 79.75 & 39.22 & 96.51 & 6.88 & 96.74
				\\
				Class 5 & 98.11 & 95.58 & 56.10 & 82.16 & 51.4 & 71.81 & 0.48 & 95.35 & 22.04 & 95.58
				\\
				Class 6 & 96.31 & 95.25 & 36.74 & 77.51 & 45.41 & 76.94 & 1.06 & 95.02 & 1.32 & 95.25
				\\
				Class 7 & 97.75 & 97.28 & 42.49 & 91.68 & 39.26 & 84.36 & 2.51 & 95.05 & 12.96 & 97.28
				\\
				Class 8 & 97.16 & 96.62 & 66.40 & 82.69 & 34.41 & 74.20 & 71.08 & 96.39 & 9.81 & 96.62
				\\
				Class 9 & 95.89 & 96.56 & 42.32 & 80.49 & 31.28 & 78.49 & 3.23 & 96.33 & 0.59 & 96.56
				\\
				\bottomrule
			\end{tabularx}
		\end{threeparttable}
	\end{table}	
	\begin{figure*}[ht]
		\centering
		\includegraphics[width=\textwidth]{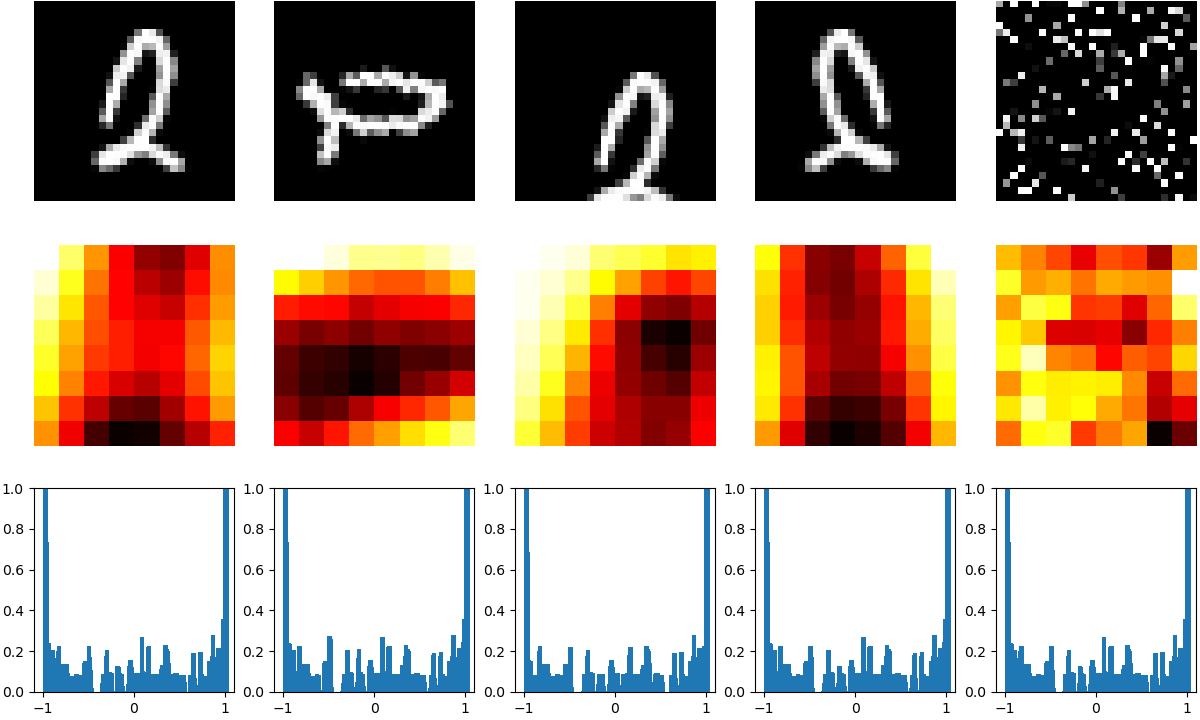}
		\caption{\centering An example of features extracted by CNN and DADM. From top to bottom, the rows are: input images, feature maps extracted from LeNet, and the histograms computed by DADM. From left to right, the columns represent: original, rotated, translated, flipped, and shuffled images. }
		\label{case_study_fig}
	\end{figure*}
	To further reveal the reason behind such robustness of our model, we also conduct a case study on the class-wise performance of DADM. The results are detailed in Table~\ref{case_study}, where the class-wise accuracy of LeNet is also listed as the baseline. While both methods demonstrate high accuracy on the unaltered images, LeNet shows relatively high sensitivity to various transformations of images across all categories, especially for digits with complex shapes or less distinct features such as ``3'' or ``6''. This sensitivity in LeNet can be attributed to its reliance on spatial features, which can be significantly altered by transformations like rotation or flipping. DADM, on the other hand, shows higher stability under these transformations across different categories. Such stability is largely due to DADM's ability to learn and utilize the global distributional information, which to some extent is invariant to spatial change. 
	\par
	Figure~\ref{case_study_fig} shows an example of the feature extraction in LeNet and DADM when classifying the digit ``2'', providing further insights. While the image is changed by various transformations, the histograms computed by DADM maintain a consistent pattern, which is in contrast to the feature maps from LeNet where each transformation results in a visibly different feature representation. This highlights that DADM can recognize the underlying distribution despite the change of spatial arrangement of pixels, and thus can robustly perform image classification against affine transformation or even more challenging transformations such as shuffling pixels.

	\section{Discussion}
	One significant advantage of this work is that it explores a new direction that more closely resembles human visual capabilities, particularly regarding handling affine transformations such as rotations. The presented method is indeed more robust under these conditions, thereby making the model more applicable to real-world scenarios where data can be in various orientations and positions.
	\par
	Furthermore, the proposed approach provides a differentiable histogram construction method for the distribution layers. This differentiability allows the distribution layers to learn not only the probability information in the raw input images, but also the distributions of features extracted by other neural network layers such as CNNs. This can further add value to distribution learning and enhance the explainability of the input feature. We will further our study to explore the abilities of our model in terms of these potential directions in future work.

	\section{Conclusion}
	In this work, we proposed differentiable arithmetic distribution learning to tackle the inherent limitations of CNNs in handling affine transformations. 
	A cornerstone of our approach was the application of a KDE-based differentiable histogram as a replacement for traditional histograms. 
	This provides us with a differentiable approach to model data distributions, thereby paving the way for seamless integration between distribution and neural networks. 
	Accordingly, we formulate a novel neural network module called DADM that effectively captures the inherent distributional attributes of the input data. 
	This results in a model that is not only robust to affine transformations, but also retains the advantages of conventional CNNs in local spatial feature extraction.
	Our experiments, which includes an ablation study and a comparison with LeNet, demonstrates the effectiveness and robustness of our approach. 
	The results show that, while the performance is comparable on original datasets, the resilience of our model against affine transformations is notably superior.

	\bibliographystyle{IEEEtran}  
	\bibliography{ref}
	
\end{document}